\documentclass[a4paper,conference]{IEEEtran}
\ifCLASSINFOpdf
\else
\fi
\usepackage{graphics}
\usepackage{float}
\usepackage{graphicx}
\usepackage{multicol, blindtext}

\begin{document}
%
\title{A Grid-based Representation for \\Human Action Recognition}

\author{\IEEEauthorblockN{Soufiane Lamghari, Guillaume-Alexandre Bilodeau}
\IEEEauthorblockA{LITIV, Dept. of Computer and Software
Engineering, \\ Polytechnique Montreal, Montreal, Canada\\
Email: soufiane.lamghari@polymtl.ca, gabilodeau@polymtl.ca}
\and
\IEEEauthorblockN{Nicolas Saunier}
\IEEEauthorblockA{Dept. of Civil, Geological and Mining Eng.,\\ Polytechnique Montreal, Montreal, Canada\\
Email: nicolas.saunier@polymtl.ca}}


%


\maketitle

\begin{abstract}

Human action recognition (HAR) in videos is a fundamental research topic in computer vision. It consists mainly in understanding actions performed by humans based on a sequence of visual observations. In recent years, HAR have witnessed significant progress, especially with the emergence of deep learning models. However, most of existing approaches for action recognition rely on information that is not always relevant for this task, and are limited in the way they fuse the  temporal information. In this paper, we propose a novel method for human action recognition that encodes efficiently the most discriminative appearance information of an action with explicit attention on representative pose features, into a new compact grid representation. Our GRAR (Grid-based Representation for Action Recognition) method is tested on several benchmark datasets demonstrating that our model can accurately recognize human actions, despite intra-class appearance variations and occlusion challenges.

\end{abstract}


%
\IEEEpeerreviewmaketitle

\section{Introduction}

Human Action Recognition (HAR) is a very popular topic in computer vision. The popularity of this task is mainly due to its use in various real-world applications such as smart video surveillance, autonomous robots, virtual reality, sport video analysis and urban planning \cite{gaur2011string,sudha2017approaches, xia2015robot,ibrahim2016hierarchical}. The goal of HAR is to identify and classify human actions from video sequences that contain spatial and temporal information related to the performed human action. Actions can be complex (e.g., preparing a meal) or simpler (e.g., walking). In this work, we focus on atomic human actions (e.g., running, dancing, jumping). Despite great progress in the last few years, human action recognition is still a challenging task due to dynamic backgrounds, occlusion, varied people appearance and imaging conditions.

In the last few years, there has been a lot of research based on deep learning to recognize human actions in videos~\cite{Karpathy2014LargeScaleVC, Simonyan2014TwoStreamCN, wang2016temporal, donahue2015long, zhou2018mict}. Since videos are 3D spatio-temporal signals, the main idea behind the majority of these studies is to extend Convolutional Neural Networks (CNNs) to include the temporal information contained in videos. Karpathy et al.~\cite{Karpathy2014LargeScaleVC} proposed several fusion techniques that slightly modify the CNN architectures to operate on stacked video frame inputs. As their results were similar to the results obtained by using individual RGB frames, these techniques were shown to not correctly model the temporal information. In order to operate in the spatio-temporal domain, Ji et al.~\cite{Ji20103DCN} proposed a 3D CNN model that performs 3D convolutions on stacked video frames to learn spatio-temporal information between consecutive frames. In addition to the fact that 3D CNNs perform similarly to 2D CNNs, they are computationally expensive to train because they contain many more parameters and do not model long range temporal information. In the same context, Simonyan et al.~\cite{Simonyan2014TwoStreamCN} proposed a two-stream CNN architecture that learns spatial appearance information from RGB frames and motion information between frames using optical flow. To improve this architecture that considers only a single frame as input, Ng et al.~\cite{Ng2015BeyondSS} and Wang et al.~\cite{Wang2016TemporalSN} proposed architectures that aggregate the convolutional features at different temporal and spatial positions. However, the streams in these two-stream CNN architectures are independent and there is no shared information between them. These architectures capture only the motion information in short time windows and do not guarantee to keep the most representative features with pooling techniques. 

Another line of research incorporates human pose sequences to represent actions as they provide valuable cues for the recognition task. Multiple studies proposed to recognize actions based on 3D poses \cite{du2015hierarchical, shahroudy2016ntu, liu2016spatio, wang2013approach}. However, these methods are less convenient for general cases, because they require special depth sensors. With the research progress in pose estimation over the past few years, some alternative approaches exploit 2D poses to recognize human actions \cite{jhuang2013towards, cheron2015p, choutas2018potion}. Similarly to the previously discussed approaches, 2D pose-based methods still represent actions by randomly leaned features. Also, they remain limited in the way they integrate the temporal information that is irrelevant with respect to the dynamic nature of human actions.

To address the aforementioned problems, we propose a novel pose-based approach for human action recognition that learns the temporal discriminative features of actions by integrating them in a compact static representation. Based on key poses, our model consists in integrating the most representative appearance features in a single grid image to obtain a more relevant representation of the performed action. In order to restrict the analysis to only the most likely information related to the action, we only consider the human region in the scene in each frame. By fusing valuable appearance features with representative poses, our grid representation mimics an explicit attention mechanism that allows us to deal with some challenges related to real-world data including occlusions and intra-class appearance variations. Furthermore, we tested our approach, called GRAR (Grid-based Representation for Action Recognition) on several datasets and found that it yields competitive or state-of-the-art results for individual actions as well as collective activities when integrated into a bottom-up setting.

The contributions of this paper are threefold: 
\begin{enumerate}
    \item We propose a new grid action representation that encodes only discriminative appearance features.
    \item We consider an explicit attention mechanism that highlights the representative poses of the action and can handle challenging situations, such as occlusions and intra-class variations.
    \item Experiments on three publicly available benchmark datasets demonstrate the effectiveness of our proposed model by achieving competitive results compared to the state-of-the-art.
\end{enumerate}

\section{Related work}
\label{related_work}
Our proposed work is related to three lines of research: deep learning-related action recognition, pose-based video action recognition and collective activity recognition in videos. In this section, we review notable studies related to these research areas and show how our proposed method differs from them.

\subsection{Deep Learning-based Action Recognition}

In recent years, deep learning methods have shown valuable capabilities in various computer vision applications ranging from image classification to action recognition in videos. CNNs \cite{lecun1998gradient} are regarded as a powerful class of models for the task of video action recognition. Simonyan et al. \cite{simonyan2014two} proposed to integrate spatial and temporal networks into a two-stream CNN architecture that is trained independently on inputs from static appearance and multi-frame dense optical flow. Similarly, based on a two-stream CNN architecture, Wang et al. \cite{wang2016temporal} introduced a model called TSN that learns video representations based on sparse temporal sampling to encode the long range temporal structure for a better understanding of the dynamics in action videos. A 3D CNN model that extends 2D convolutions to 3D convolutions is proposed by Ji et al. \cite{Ji20103DCN} to learn spatio-temporal information between stacked consecutive frames. The problem of 3D CNN is that they contain many more parameters, which makes them computationally expensive. Moreover they do not allow modeling of long range temporal information.

Another widely adopted approach in this context is the use of Recurrent Neural Network (RNN) and its variants (e.g., LSTM and GRU). These networks have demonstrated an impressive performance in modeling long-term dependencies between frames. Donahue et al. \cite{donahue2015long} applied a long-term recurrent convolutional network to model visual time-series to recognize actions. In a different work, Du et al. \cite{du2017rpan} introduced a recurrent network based on pose and attention mechanisms, where the spatio-temporal evolution of the human pose is used to guide the process of recognizing human actions in videos. Recently, Li et al. \cite{li2018videolstm} proposed an end-to-end sequence learning framework for action classification that integrates attention via a convolutional LSTM network.

\subsection{Pose-based Video Action Recognition}

Human pose is considered as an appearance clue that can be leveraged to guide the process of action recognition in videos. In this context, Wang et al. \cite{wang2013approach} proposed to infer the best poses for each frame by extracting spatial-part-sets and temporal-part-sets using a contrast mining algorithm \cite{dong1999efficient}, where the output is then fed to a SVM classifier in order to recognize human actions in videos. Nie et al. \cite{xiaohan2015joint} proposed a similar approach based on a spatial-temporal And-Or graph hierarchical model that decomposes human actions into three levels including poses, spatio-temporal parts and body joints. Later, Zolfaghari et al. \cite{zolfaghari2017chained} integrate poses, motions, and raw images into a three-stream architecture to improve the action recognition performance. Recently, Choutas et al. \cite{choutas2018potion} proposed an approach called PoTion that jointly encodes appearance and motion of semantic keypoints into a clip-level representation serving as input for a shallow CNN.

\subsection{Collective Activity Recognition}

The past few years have witnessed an increasing interest by the research community about collective activity recognition \cite{ibrahim2016hierarchical},\cite{shu2017cern}, \cite{bagautdinov2017social}, \cite{qi2018stagnet},  \cite{biswas2018structural}. A notable work was introduced by Choi et al. \cite{choi2009they} where they described the activity of a person based on spatio-temporal descriptors in order to infer the high-level collective activity. Recently, multiple deep learning based models have been proposed in this context. A two-stage hierarchical temporal model was introduced by Ibrahim et al. \cite{ibrahim2016hierarchical} to recognize collective activities. In the first stage, they analyze the temporal dynamics of each person with an LSTM network. Then, they aggregate this information in the second stage with the encoded temporal group dynamic, in order to learn interactions between people that contributes to recognize collective activities. On the other hand, Deng et al. \cite{deng2016structure} proposed a framework that combines graphical models and deep neural networks. In this model, nodes are representing both people and the scene, which allows message passing between outputs. Later, StagNet was proposed by Qi et al. \cite{qi2018stagnet} where a semantic graph is used to model individual actions as well as their corresponding spatial relations, whereas the temporal interactions are modeled with a structural-RNN architecture.

\begin{figure*}[ht]
\centering
\includegraphics[scale=0.36]{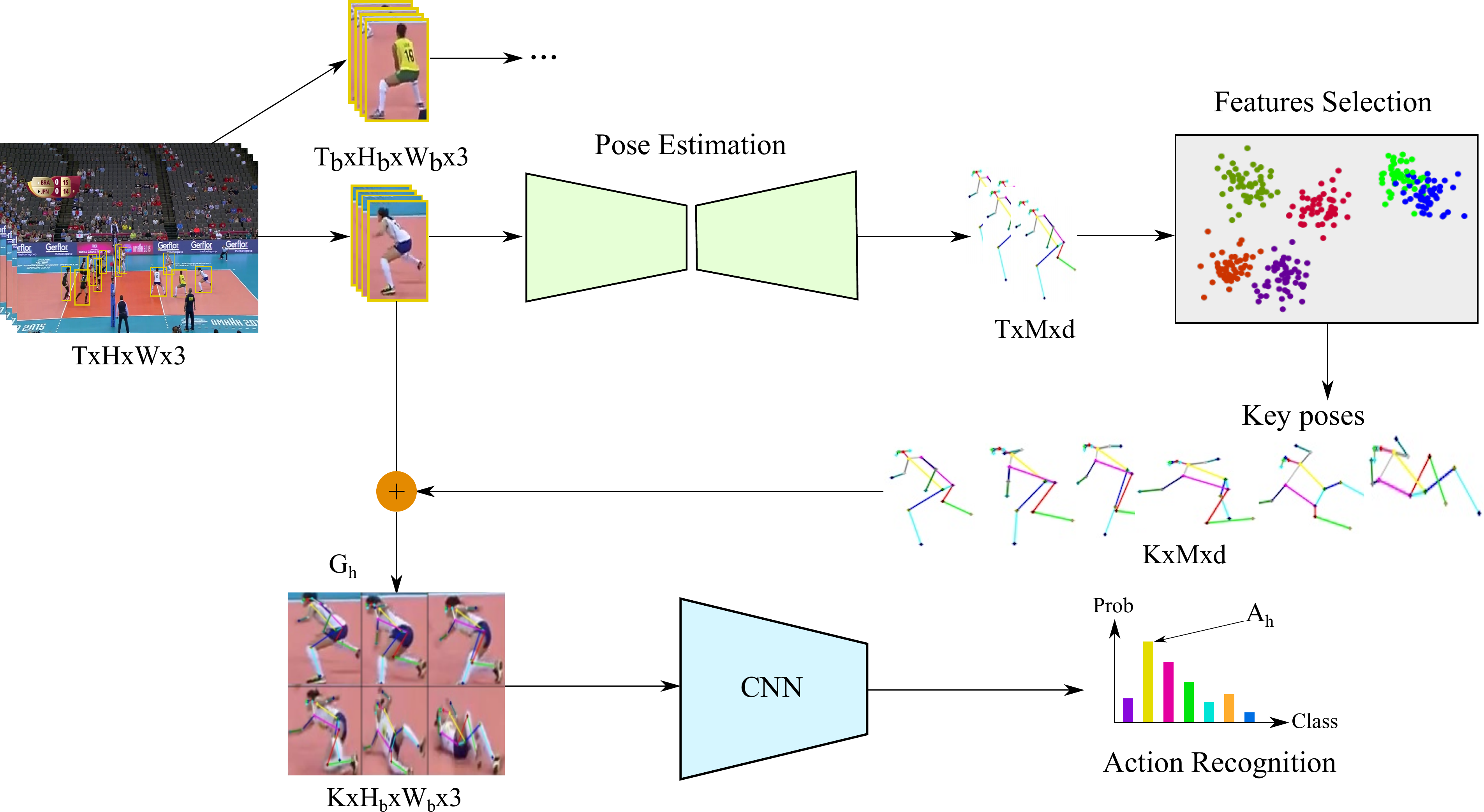}
\caption{The pipeline of our proposed GRAR model. From each frame, we extract human poses using a pose estimation method. We select then the most representative key poses for each human based on an unsupervised clustering method. These estimated key poses are combined with their corresponding RGB sources and then concatenated. This results in a compact grid representation $G_{h}$, that encodes the relevant RGB and pose information related to the performed action. Finally, we train a deep convolutional neural network on the obtained grid representations to predict the corresponding action category $A_{h}$.}
    \label{pipeline}
\end{figure*}

Different from these methods, we propose a novel action recognition approach that relies mainly on relevant poses to build an action representation instead of considering random features or frames. Additionally, with an attention guided by key poses, our approach is more robust to occlusion and intra-class appearance variation problems. Following a bottom-up design, our approach can be successfully leveraged to recognize activities based on individual actions.

\section{Grid-based Representation For Human Action Recognition}
\label{proposed_approach}

In this section, we present the overall design of our novel GRAR model, whose ultimate goal is to recognize efficiently the actions of all persons in a video sequence, by combining important temporal features associated with poses. The overall pipeline is illustrated in Fig. \ref{pipeline}.

To obtain a more relevant representation of the performed action, first of all, we select the human region in the scene in each frame, instead of taking the entire frame. This restricts the analysis to focus only on the most relevant information related to the person's action. Given the tracks of every person, we extract the sequence of the 2D human poses at each time step. Once the poses are extracted, we normalize the joint coordinates with respect to the bounding box position at each time $t$ to get their corresponding relative positions. Based on these normalized joint positions, we select the most representative poses of each action using an unsupervised clustering algorithm. This will allow us to keep only the information of interest about the performed action. We will refer to these representative poses as key poses in the rest of the paper.

From the chosen key poses, we create our new static grid representation that integrates only the most discriminative temporal RGB and pose information, which allows our method to deal with periodicity in actions as well as occlusions and intra-class variations problems. Indeed, since we are using key poses, frames where the person is too much occluded will likely be ignored because they will result in unstructured or infrequent poses. Other times, poses can allow to extract information through occlusions (see Fig.~\ref{occlusions}). Finally, we recognize the performed actions by training a convolutional neural network on sets of these grid images. Because actions are represented in a grid image, we can benefit from pre-trained image classification networks.

In the following sections, we describe in details the main components of our proposed model. Since we are interested mainly in human action recognition in this work, we assume that humans are already detected.

\subsection{Human Pose Estimation}

Human actions are highly correlated with their corresponding poses. A 2D human pose is a continuous representation of the body parts in the image space. Let $V_{h} = \{V_1,V_2,...,V_t,...,V_T\} \in {\rm I\!R}^{T \times M \times d}$ denote the pose sequence for a human $h$ in $T$ frames, where $d$ is the spatial dimensions and $M$ is the total number of human body keypoints (joints). The pose vector $V_t$ is defined as $V_t = \{w_{t1}, w_{t2},...,w_{tm},...,w_{tM}\}$ where $w_{tm}$ is the spatial coordinate of the keypoint $m$ in a single frame at time $t$. In this work, we use the recently published HighResolution Net (HRNet) pose estimation model~\cite{wang2020deep} for its good performance on small persons. HRNet is a top-down architecture based on repeated multi-scale fusions between parallel multi-resolution sub-networks. For computational purposes, in our proposed method, we use the HRNet-W32 network trained using the $MSE$ loss function defined as:
\begin{equation}
    \mathcal{L}_{mse} = \frac{1}{M} \sum_{m=1}^{M} || C_m - \hat{C_m}||_{2}^{2}
\end{equation}

where $\hat{C_m}$ and $C_m$ are the predicted and the ground truth confidence map for the $m^{th}$ joint respectively.

Given a sequence of video frames $S \in {\rm I\!R}^{T \times H \times W \times 3}$, the model predicts the pose sequence $V_h$ for a human $h$ based on the bounding box sequence $U_h \in {\rm I\!R}^{T_h \times H_h \times W_h \times 3}$. As we use the COCO keypoints format \cite{lin2014microsoft} for the 2D pose estimation, we consider $M=17$ and $d=2$. In some special cases, object detection and tracking methods can fail to estimate good quality bounding boxes. In the case of humans, this corresponds to a bounding box that does not cover all the body joints of the person of interest. To solve this issue, once we get the pose keypoint coordinates, we apply a bounding box refinement process, which consists in modifying the bounding box coordinates $U_{h}^{'} \in {\rm I\!R}^{T_h \times H_h^{'} \times W_h^{'} \times 3}$ to enclose the extreme joints coordinates.

\subsection{Relevant Features Selection}

Frame selection is a challenging task in action recognition in RGB videos. Considering all the frames or choosing some frames randomly to represent an action induce redundant or irrelevant information in the learning process, which is directly reflected on the final classification accuracy. In order to reduce the complexity of our proposed model and increase its generalization ability, we propose to focus on the most relevant information by extracting key poses, a subset of distinctive human poses for each action. For example, the distinctive poses for the action "running" can match frames where the right hand, and both the left knee and foot are all heading forward, in opposite directions to the left hand and the right knee and foot. At first, we make the keypoint sequence $V_h$ invariant to the position in the scene and the scale of the person of interest. Based on the corrected bounding box sequence $U{'}_{h}$, we transform the coordinates of each vector $V_t$ from the frame $t$ coordinate space to the corresponding bounding box coordinate space $U^{'}_{t}$. Then, we normalize the obtained coordinates with respect to the dimensions of the considered bounding boxes.

Now that we constructed the $V^{'}_{h} \in {\rm I\!R}^{T \times M \times d}$ sequence for each person in the same normalized coordinate space, we proceed to cluster these keypoint sequences in order to extract the most discriminative poses, that are the key poses. To this end, we employ the well-known Partitioning Around Medoids (PAM) clustering algorithm~\cite{kaufman1987clustering} based on a pairwise dissimilarity metric. This method is shown to be more robust to outliers than the sum of squared Euclidean distances used by K-Means. This fact is also demonstrated by our experiments in Section~\ref{experiments}. 

Formally, given an action represented by the $V^{'}_{h}$ pose sequence, PAM provides us with a set of $K$ pose clusters $(C_1, C_2...C_K)$ along with their reference pose medoids  $V_{h}^{*} = \{V_{k_{1}},V_{k_{2}},...,V_{k_{K}}\} \subset V^{'}_{h}$ and $V_{h}^{*} \in {\rm I\!R}^{K \times M \times d}$ (i.e., the most centrally located pose in a cluster), where both the intra-cluster poses similarity and the inter-cluster dissimilarity are maximized. The learning process of extracting key poses for each person performing a specific action is done as follows: at first, we randomly select a set of $K$ poses for each person, then we assign the remaining poses to the cluster with the most similar medoid to them. After that, we select an arbitrary non medoid pose $x$ and compute the cost of swapping the initial medoid with the new candidate medoid. The updated total cost is based on the L1 norm and is defined as:
\begin{equation}
    V_{k_{1}},V_{k_{2}},...,V_{k_{K}} =  argmin \sum_{i=1}^{K} \sum_{x \in C_{i}} || x - V_{k_{i}}||_{1}
\end{equation}

After convergence, the obtained poses medoids serve as the key poses, which will be used as the input feature vector for the next module.

\subsection{Grid Representation Learning}

Several research works in the literature have explored multiple fusion techniques to integrate temporal information \cite{Karpathy2014LargeScaleVC} and complementary modalities \cite{Simonyan2014TwoStreamCN} in order to improve the final action recognition accuracy. However, these methods remain limited in the way they integrate information in time. In addition, they rely on randomly selected information which results in less representative features. Different from these methods, our GRAR model is based only on relevant RGB information. Without requiring any additional annotations (i.e, pose, skeleton data), the estimated key poses are used as an explicit attention mechanism. By fusing temporal RGB and pose features into a grid image representation, 
our model efficiently encodes key patterns needed to recognize human actions. 

\begin{figure}[ht]
    \centering
    \includegraphics[scale=0.4]{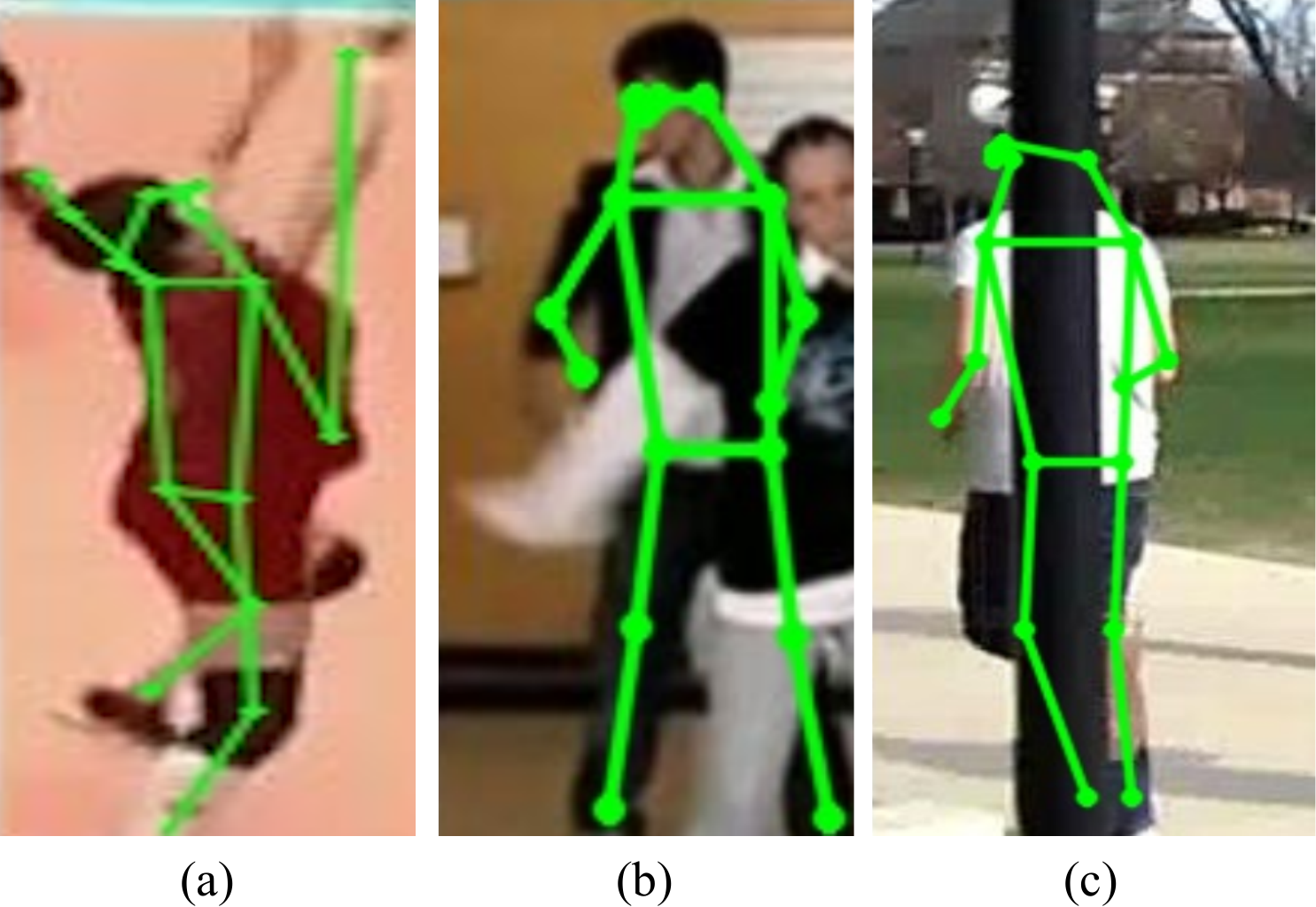}
    \caption{Examples from the Volleyball and CAE datasets of (a) a pose estimation failure case considered as outlier during key poses selection. (b) and (c) show occluded humans handled with our explicit attention on pose.}
    \label{occlusions}
\end{figure}

In order to create our new grid structure, we proceed as follows: For each key pose in $V_{h}^{*}$, we get the RGB information of interest $I_{{h}_k}$ based on its corresponding bounding box. In an explicit manner, we put attention on the key poses by modifying the pixels $p \subset I_{{h}_k}$ if $p \in v_{h_k}$ so that the pose vector is encoded directly in the selected relevant RGB region.  This attention-guided by pose technique allows us to deal with challenging real-word situations such as intra-class variations and occlusions. Fig. \ref{occlusions} (b) shows an example where our model can successfully compensate missing information caused by occlusions.

\begin{figure}[ht]
    \centering
    \includegraphics[scale=0.4]{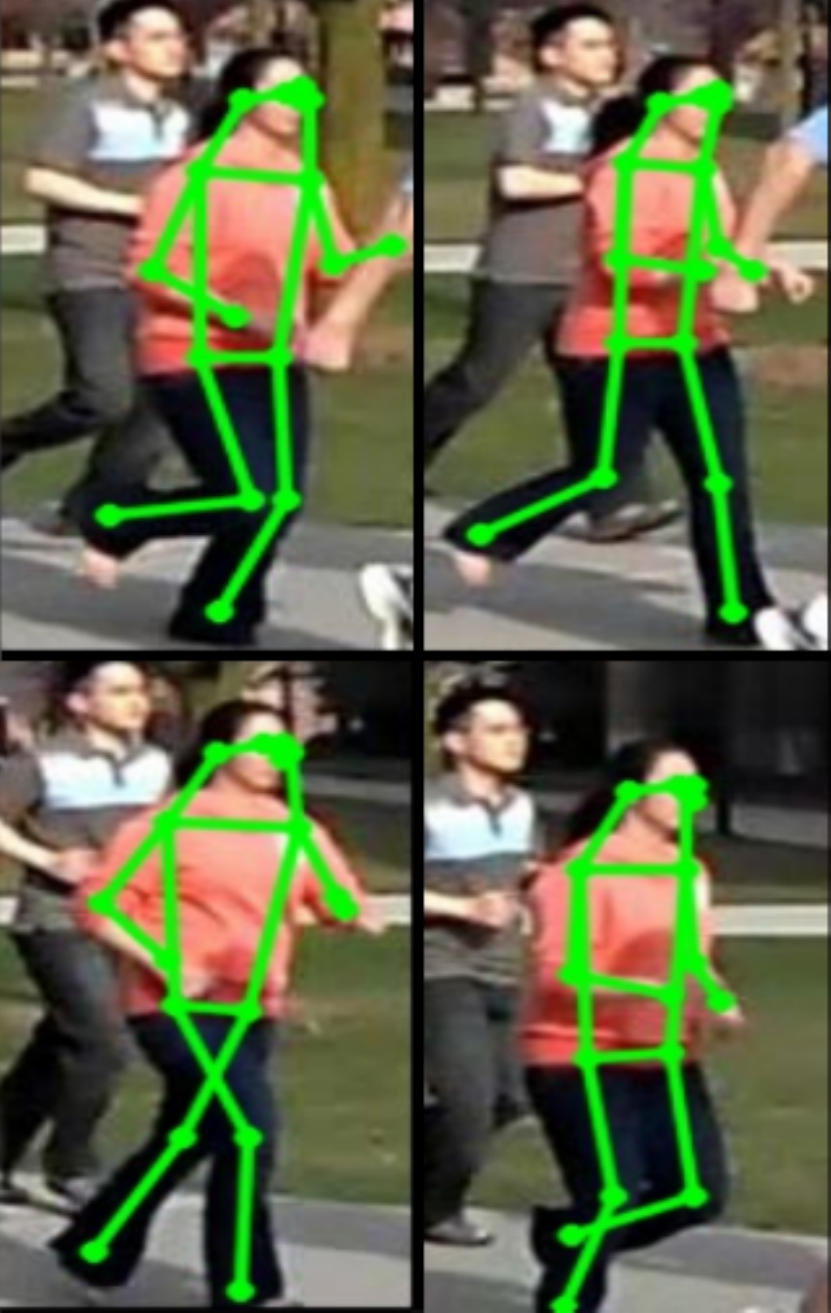}
    \caption{Example of a grid image from the CAE dataset.}
    \label{grid}
\end{figure}

Afterwards, we put each fused $I_{{h}_k}$ in a separated cell forming our main grid structure $G_{h}$ (see Fig. \ref{grid} for an example of a grid image). Before grouping the obtained cells, we concatenate each cell with a zeros-valued border. Such operation is necessary to avoid learning unnecessary patterns created by adjacent cells, when the convolution kernel sizes are larger than the inter-cells spacing. Our experiments show that considering a 3 pixel-wide boundary is enough according to the filter sizes of the adopted CNN architecture. It is to note that all of the $I_{{h}_k}$ cells are not re-scaled, keeping them at their original resolution allows our model to learn jointly the features, as well as their corresponding distance from the camera. 

Once the grids are created, we train a CNN to learn human action patterns from these sets of grids. We adopt the Inception-ResNet-v2 model \cite{szegedy2017inception} pre-trained on the ImageNet dataset \cite{deng2009imagenet} as our backbone CNN architecture. This model is a hybrid Inception network that uses residual connections rather than filter concatenation. We use the categorical cross entropy as our loss function which is given by:

\begin{equation}
    \mathcal{L}_{CE} = - \frac{1}{N} \sum_{i=1}^{N} \log \frac{e^{w^{T}_{y_i}G_i+b_{y_i}}}{\sum_{j=1}^{n} e^{w^{T}_{j}G_i+b_{j}}}
\end{equation}

where $G_i$ refers to the $i^{th}$ grid representation, $N$ is the number of training grids, $y_i$ is the class label of $G_i$, W is the learned weight matrix and $b$ is the intercept.

Considering only the human region in the scene instead of the whole frame information, allows us to combine a multitude of cues within the same image representation, without reducing excessively the original image resolution. Excessive downsampling of the image resolution to fall within the CNN input size, often results in loosing information valuable for action recognition. Moreover, being based solely on the human region with pose information maximizes the information closely related to the performed actions, and thus allows our model to generalize well to other scenes. For example, in case of a walking action, if our model is trained only with samples of humans walking on a sidewalk, in the inference phase it will be able to recognize that a human walks even if it is on the grass. This can be explained by the fact that the extracted deep features are more related to the human itself than to his environment.

\section{Experiments}
\label{experiments}

In this section, we evaluate the performance of our proposed model. We first present the considered benchmark datasets and the implementation details. Then we report the results of a series of ablation studies to analyze the impact of each component of our GRAR model on the recognition performance, followed by a comparison with the state-of-the-art.

\subsection{Datasets}

We evaluate our model on three publicly available datasets: the Collective Activity dataset~\cite{choi2009they}, the Collective Activity Extended dataset~\cite{choi2011learning} and the Volleyball dataset~\cite{ibrahim2016hierarchical}.

\subsubsection{Collective Activity dataset (CA) \cite{choi2009they}} The Collective Activity dataset is a popular dataset for both individual action and group activity recognition. It contains 44 video sequences with a resolution of 640$\times$480 pixels from 5 individual actions and group activity categories (talking, crossing, queuing, waiting, walking). The collective activity label of a scene is defined based on the performed action of the majority of individuals. For train/test split, we follow the same evaluation protocol suggested in~\cite{choi2009they}.

\subsubsection{Collective Activity Extended dataset (CAE)  \cite{choi2011learning}} The Collective Activity Extended dataset is an extended version of the original Collective Activity dataset where the "Walking" activity was replaced by two new activities "Jogging" and "Dancing". The reason why "Walking" was removed is because in some scenarios this activity is mixed with the Crossing activity.  To train our model, we followed the testing scheme mentioned in~\cite{deng2016structure} and use $\frac{2}{3}$ of the videos for training and the rest for testing.

\subsubsection{Volleyball dataset \cite{ibrahim2016hierarchical}} The Volleyball dataset contains 4830 frames collected from 55 YouTube videos and are all about Volleyball games. Each player is labeled with one of these actions: moving, spiking, waiting, blocking, jumping, setting, falling, digging and standing. We adopted the same testing setup used in~\cite{ibrahim2016hierarchical}, where $\frac{2}{3}$ of the data is used for training and $\frac{1}{3}$ for testing.

\subsection{Implementation Details}

We implemented our proposed model using the TensorFlow library~\cite{abadi2016tensorflow}. We use
Inception-ResNet-v2~\cite{szegedy2017inception} as our backbone CNN architecture pre-trained on the ImageNet dataset~\cite{deng2009imagenet}. This network consists in 164 layers, with an image input size of 299$\times$299. For all the experiments and datasets, we utilized stochastic gradient descent with ADAM \cite{kingma2014adam} and set the optimizer hyperparameters to $\beta_1$ = 0.9, $\beta_2$ = 0.999, $\epsilon$ = 0.001. For the CA and CAE datasets, we used  4 key poses and trained the model in 100 epochs with a minibatch size of 16 and an initial learning rate starting from $10^{-3}$ then reduced with a factor of 0.2 after 10 patience epochs. For the Volleyball dataset, we used 6 key poses and trained the network with a learning rate of $10^{-5}$ for 130 epochs with a mini batch size of 32.

To track humans in the scene, for the Volleyball dataset, we used the tracker proposed by Cao et al. \cite{7410695}, which is implemented in the Dlib library \cite{lan2011discriminative}. For the CA and CAE datasets, we used the tracklets provided by \cite{choi2009they}. We used the HighResolution Net (HRNet) algorithm \cite{wang2020deep} to compute human postures across frames. Specifically, we used the pose-hrnet-w32 architecture trained on COCO dataset. All our experiments were run on a single TITAN Xp NVIDIA GPU.

\subsection{Ablation Studies}

In order to explore the effect of every component of our model on the performance, we conducted extensive ablation studies on the CAE dataset with the following variants.

\paragraph*{Clustering Analysis}

We start by studying the impact of different clustering settings on the recognition performance. The goal here is to find key poses forming the discriminative grid representation. We compare three clustering methods, namely, PAM, K-means and Gaussian Mixture Model (GMM) estimated with the Expectation-Maximization algorithm. For action representation learning, we used the same CNN architecture and parameters settings. The results are reported in table \ref{Tab0}.

\begin{table}[ht]
\caption{Impact of the clustering algorithms and number of key poses on the performance of our model on the CAE dataset. \textbf{Boldface: Best result.}}
\label{Tab0}
\centering
\begin{tabular}{l c c c}
\cline{2-4}
\cline{2-4}
    & K-means & PAM & GMM \\ \hline
K = 2 &   94.0\%      &                 94.6\%             &          90.3\%              \\
K = 4 &   94.5\%      &          \textbf{95.2}\%                    &            91.7\%            \\
K = 6 &     92.6\%    &                  93.9\%            &           91.9\%         \\  \hline   
\end{tabular}
\end{table}

Compared with K-means and GMM, we can see that PAM gives the highest accuracy 95.2\%. This can be explained by the fact that PAM is robust against outliers. In fact, it minimizes the average dissimilarity of human poses in each cluster, rather than minimizing the squared sum of each intra-cluster as adopted by K-means. It is important to highlight that the poses that we use in our model are not manually labeled but instead, are predicted with the HRNet model that can fail sometimes to estimate high quality poses. Fig. \ref{occlusions} (a) illustrates an example of a failure case in pose estimation, which is considered in our study as an outlier pose. Compared to PAM and K-means, GMM achieves the worst recognition rate in this experiment. This can be explained by the fact that the nature of our input pose data is not normally distributed.

Additionally, we evaluated the effect of the number of key poses on the recognition performance. In this experiment, we consider several numbers of key poses, starting from two to six. Intuitively, one can say that the more key poses we consider, the more we gain information about the performed action. However, our evaluation on the datasets demonstrates that above a certain number of key poses, the recognition accuracy starts to decrease. We found that the performance is directly related to the input size of the backbone CNN model that we are using. In fact, by downsampling images to fit the input size of the CNN, some important cues for the recognition task are usually lost. So far, our findings demonstrate that larger downsampling rate makes performance poorer. After extensive experiments, we concluded that the number of key poses must be chosen based on the average size of humans in the scene. Furthermore, to keep the original full resolution for each key pose image, our grid must have a resolution that is close enough to the default CNN input size.

\paragraph*{Features selection}

Next, we study the importance of features selection for human action recognition. We compared random selection against pose-based selection for grid representation. For pose-based selection, we also compared two strategies, using only the poses features (K-Pose, in this case, key poses are drawn in white over a black background) and using only the RGB image corresponding to a key pose (K-RGB). For action representation learning, we employed the same CNN architecture in the three experiments. The results are reported in table \ref{Tab1}.

Compared with random selection, choosing RGB information based on human pose (K-RGB) gives an improvement of 3.1\%. This demonstrates that using the pose estimation to select the RGB data provides us with relevant and discriminative information for human action recognition. On the other hand, using only pose features (K-Pose) with a standard CNN model do not yield good results. Such experience reveals that despite the pose containing valuable cues about the performed action, using it solely does not provide the standard 2D CNN with enough significant features for recognition.

\begin{table}[ht]
\caption{Impact of different modules on the accuracy of GRAR based on CAE dataset. \textbf{Boldface: Best result.}}
\label{Tab1}
\centering
\resizebox{0.5\textwidth}{!}{\begin{tabular}{l c}
\hline
Model Variants & Accuracy\\
\hline
Random                       &              89.2\%                 \\
Key poses only (K-Pose)      &                80.5\%               \\
Key Frame (K-RGB)            &                          92.3\%     \\
Key Frame+Box enhancement (K-RGB+EB) &              92.9\%       \\
Key Frame+Box enhancement+Pose Attention (K-RGB+EB+PA) & \textbf{95.2}\% \\
\hline
\end{tabular}
}
\end{table}

\paragraph*{Bounding Boxes Enhancement (EB)}

Here, we evaluate the effectiveness of making use of the human pose to correct inaccurate bounding boxes used in GRAR pipeline. As illustrated in Table \ref{Tab1} at the K-RGB and K-RGB+EB rows, we can see that correcting the human bounding boxes give us 0.6\% of improvement. This emphasizes the importance of incorporating human joint features to enhance bounding boxes quality, which is useful not only for the action recognition task, but could be useful to other computer vision problems.

\paragraph*{Pose Attention (PA)}

Finally, we study the impact of the introduced pose attention technique, where a key pose is drawn over the corresponding RGB image. We compared the recognition performance of K-RGB+EB against using K-RGB+EB combined with pose-based attention. In both experiments, we considered the enhanced version of the bounding boxes (EB). As indicated in Table \ref{Tab1} at the K-RGB+EB and K-RGB+EB+PA rows, we can conclude that putting explicit pose-attention on appearance representation improves the recognition performance by around 2.3\%. This indicates that if the estimated pose is of good quality, an attention mechanism based on it makes the model more robust against intra-class variations and occlusions problems, as explained in Fig. \ref{occlusions}.

\subsection{Comparison with the State-of-the-Art}

In this section, we compare the performance of our GRAR model with respect to several state-of-the-art methods including Learning context \cite{choi2011learning}, Social Cues for activity recognition \cite{tran2013social}, Hierarchical Deep Temporal Model \cite{ibrahim2016hierarchical}, Structure Inference Machines \cite{deng2016structure}, CERN \cite{shu2017cern}, StagNet \cite{qi2018stagnet}, Fast collective activity \cite{zhang2019fast_art}, Gaim \cite{lu2019gaim}, ARG \cite{wu2019learning}, SSU \cite{bagautdinov2017social} and SRNN \cite{biswas2018structural}.

\subsubsection{Results on the Collective Activity dataset}

Now that we evaluated multiple variants of our model for individual action recognition, our goal here is to explore the ability of our model to recognize collective activities based on the individual ones. As previously done, we derive the collective activity label of a scene based on the performed action of the majority of individuals. Moreover, we do not use any ground truth annotations about the pose.

Table \ref{cad} summarizes the state-of-the-art performance on the Collective Activity dataset (CA). Our model with the pose-based grid representation outperforms the compared state-of-the-art methods. For example, our model achieves $\approx$10\% higher accuracy than recent methods based on hierarchical relational networks \cite{ibrahim2016hierarchical} and recurrent neural networks for activity recognition \cite{deng2016structure}. This is mostly because we focus primarily on highly discriminative RGB features along with their corresponding poses.

\begin{table}[!t]
\caption{Comparison of the activity recognition performance of state-of-the-art methods versus our model evaluated on CA dataset. \textbf{Boldface: Best result.}}
\label{cad}
\centering
\begin{tabular}{l c}
\hline
Method & Accuracy\\
\hline
Choi et al. \cite{choi2011learning} & 70.9\% \\
Tran et al. \cite{tran2013social} & 78.7\% \\
Ibrahim et al. \cite{ibrahim2016hierarchical}   &    81.5\% \\
Deng et al. \cite{deng2016structure} & 81.2\% \\
Shu et al. \cite{shu2017cern}   & 87.2\%\\
Qi et al. \cite{qi2018stagnet} & 89.1\% \\
Zhang et al. \cite{zhang2019fast_art} & 83.8\% \\
Lu et al. \cite{lu2019gaim} & 90.6\% \\
Wu et al. \cite{wu2019learning} & 91.0\% \\
\hline

GRAR (Ours) & \textbf{91.5}\% \\

\hline
\end{tabular}
\end{table}

\begin{table}[ht]
\caption{Comparison of the activity recognition performance of state-of-the-art methods versus our model evaluated on CAE dataset. \textbf{Boldface: Best result.}}
\label{cade}
\centering
\begin{tabular}{l c}
\hline
Method & Accuracy\\
\hline

Choi et al. \cite{choi2011learning} & 82.0\% \\

Tran et al. \cite{tran2013social} & 80.7\% \\

Ibrahim et al. \cite{ibrahim2016hierarchical}   &   94.2\% \\

Deng et al. \cite{deng2016structure} & 90.2\% \\

Qi et al. \cite{qi2018stagnet} & 89.7\% \\

Lu et al. \cite{lu2019gaim} & 91.2\% \\

Zhang et al. \cite{zhang2019fast_art} & 96.2\% \\

\hline

GRAR (Ours) & \textbf{97.4}\% \\

\hline
\end{tabular}
\end{table}

\subsubsection{Results on the Collective Activity Extended dataset}

The experimental results of human activity recognition on the CAE dataset are shown in Table \ref{cade}. Our pose-based grid model again achieves the state-of-the-art performance with 97.4\% collective activity recognition accuracy. This performance demonstrates the effectiveness of choosing relevant RGB information and incorporating key pose features as an explicit attention mechanism in compensating the model weakness when facing ambiguous RGB appearances.

\subsubsection{Results on the Volleyball dataset}

We further conducted experiments on the Volleyball dataset. Table \ref{volley} shows the comparison of our proposed model with different recent state-of-the-art methods for individual action recognition. 
As can be seen, our GRAR model outperforms most of the state-of-the-art methods \cite{ibrahim2016hierarchical, shu2017cern, bagautdinov2017social, qi2018stagnet, biswas2018structural} with an accuracy of 82.9\%. It is also highly competitive to the ARG method \cite{wu2019learning}, mostly because this latter uses a graph convolutional network that encodes complex actor relation.






\begin{table}[!t]
\caption{Evaluation of action recognition performance of state-of-the-art methods versus our proposed model on the Volleyball dataset. \textbf{Boldface: Best result.}}
\label{volley}
\centering
\begin{tabular}{l c}
\hline
Method & Accuracy\\
\hline

Ibrahim et al. \cite{ibrahim2016hierarchical}   &    75.9\% \\
Shu et al. \cite{shu2017cern}   & 69.0\%\\
Bagautdinov et al. \cite{bagautdinov2017social} & 82.4\% \\
Qi et al. \cite{qi2018stagnet} & 81.9\% \\
Biswas et al. \cite{biswas2018structural} & 76.6\% \\

Wu et al. \cite{wu2019learning} & \textbf{83.1}\%\\


\hline

GRAR (Ours) & 82.9\% \\

\hline
\end{tabular}
\end{table}

\section{Conclusion}
\label{conclusion}

In this paper, we have presented GRAR, a novel pose-based model for human action recognition that uses a grid image of key poses. Our results consistently demonstrate that selecting RGB appearance based on the most discriminative human poses and combining them together in an image leads to considerable improvements. We obtained promising results compared to state-of-the-art approaches on three public benchmark datasets. Our proposed method has several benefits: 1) it is compact, 2) it exploits powerful CNN architectures designed for image classification tasks without requiring any architectural changes, and 3) it is robust against occlusions, intra-class action variations and incorrect human poses estimation.

\section*{Acknowledgment}
This work was supported by the National Sciences and Engineering Research Council of Canada (NSERC). We thank NVIDIA Corporation for their donation of a Titan Xp GPU card.




\bibliographystyle{IEEEtran}
\bibliography{IEEEexample.bib}
%



\end{document}